\documentclass{article}

\usepackage{microtype}
\usepackage{graphicx}
\usepackage{subfigure}
\usepackage{booktabs} 

\usepackage{hyperref}

\usepackage[accepted]{icml2020}

\icmltitlerunning{Backdoor attacks and defenses in feature-partitioned collaborative learning}

\def\bx{\mathbf x}

\def\mL{\mathcal L}
\def\mD{\mathcal D}

\def\btheta{\mathbf \theta}
\def\bTheta{\mathbf \Theta}

\usepackage{amsmath}
\usepackage{accents}

\begin{document}

\twocolumn[
\icmltitle{Backdoor attacks and defenses in feature-partitioned collaborative learning}



\icmlsetsymbol{equal}{*}

\begin{icmlauthorlist}
\icmlauthor{Yang Liu}{we}
\icmlauthor{Zhihao Yi}{we}
\icmlauthor{Tianjian Chen}{we}
\end{icmlauthorlist}

\icmlaffiliation{we}{Webank, Shenzhen, China}

\icmlcorrespondingauthor{Yang Liu}{yangliu@webank.com}
\icmlcorrespondingauthor{Zhihao Yi}{zhihaoyi@webank.com}

\icmlkeywords{feature-partitioned collaborative Learning, backdoor attack, defense}

\vskip 0.3in
]



\printAffiliationsAndNotice{\icmlEqualContribution} 

\begin{abstract}
Since there are multiple parties in collaborative learning, malicious parties might manipulate the learning process for their own purposes through backdoor attacks. However, most of existing works only consider the federated learning scenario where data are partitioned by samples. The feature-partitioned learning can be another important scenario since in many real world applications, features are often distributed across different parties. Attacks and defenses in such scenario are especially challenging when the attackers have no labels and the defenders are not able to access the data and model parameters of other participants. In this paper, we show that even parties with no access to labels can successfully inject backdoor attacks, achieving high accuracy on both main and backdoor tasks. Next, we introduce several defense techniques, demonstrating that the backdoor can be successfully blocked by a combination of these techniques without hurting main task accuracy. To the best of our knowledge, this is the first systematical study to deal with backdoor attacks in the feature-partitioned collaborative learning framework.
\end{abstract}

\section{Introduction}
Federated Learning (FL) \cite{DBLP:journals/corr/McMahanMRA16,Yang-et-al:2019} is a collaborative learning framework for training deep learning models while protecting data privacy. In \cite{DBLP:journals/corr/McMahanMRA16}, data samples are distributed by different participants thus it can be regarded as sample-partitioned collaborative learning. However, feature-partitioned collaborative learning \cite{Yang-et-al:2019, Hu2019FDMLAC,liu2019communication} can be another important scenario in many real world applications. For example, a bank may improve its credit risk model by collaborating with an E-commerce company having a large overlap of users. In this case, only the bank has the labels and both parties have different sets of features. 

Many recent works \cite{Bagdasaryan2018backdoor,pmlr-v97-bhagoji19a,xie2020dba} have shown that unprotected federated learning is vulnerable to model poisoning. The attacker manipulates the model's performance on an attacker-chosen backdoor task while maintaining the performance of the main task. For example, the attacker can choose images with some specified features, then manipulate the model to misclassify these images into an attacker-chosen category. In feature-partitioned collaborative learning, it is not clear yet whether similar attacks can be performed since parties do not have complete sets of features and labels. And in the case that backdoor attacks succeed, it is not yet clear how to defend these attacks. 

\section{Contributions}
In this paper, we show that backdoor attacks can be introduced successfully to the feature-partitioned collaborative learning by modifying the messages exchanged between different parties, even at parties with no access to training labels. The backdoor tasks achieve high accuracy on both backdoor and main tasks. To defend such backdoor attacks, we evaluated three defense techniques: adding trainable layers to active party, adding noise and gradient  sparsification. Our comprehensive experimental evaluations demonstrate that backdoor attacks can be blocked by a combination of these techniques.

\section{Related Work}
\cite{Bagdasaryan2018backdoor} introduced a backdoor attack to federated learning by replacing the global model with a targeted poisoning model and discussed effectiveness of defense strategies including differential privacy and anomaly detection. \cite{pmlr-v97-bhagoji19a} carried out stealthy model poisoning attack to federated learning by alternatively optimizing for stealth and the adversarial objectives. \cite{xie2020dba} introduced distributed backdoor attacks to federated learning to decomposes a global attack into several local ones and demonstrated superior attack performance. \cite{Sun2019CanYR} studied backdoor and defense strategies in federated learning and show that norm clipping and “weak” differential privacy mitigate the attacks without hurting the overall performance. All the works above consider the sample (horizontally)-partitioned scenario. 

Feature-partitioned collaborative learning frameworks have been developed for models
including trees~\cite{SecureBoost}, linear and logistic regression
\cite{DBLP:journals/corr/abs-1902-04885,Hardy2017PrivateFL,Gratton2018,Kikuchi2017PrivacyPreservingML,HuADMM}, and neural networks \cite{liu2018ftl,Hu2019FDMLAC}.
Privacy-preserving techniques such as Homomorphic Encryption (HE)
\cite{Rivest1978,Acar:2018:SHE:3236632.3214303}, Secure Multi-party Computation (SMPC)
\cite{Yao:1982:PSC:1382436.1382751} and Differential Privacy (DP)
\cite{Dwork:2006:DP:2097282.2097284} are typically applied in these frameworks to
preserve user privacy and data confidentiality. To the best of our knowledge, model
poisoning or backdoor attacks in such a scenario have not been studied systematically. 

\section{A Feature-partitioned collaborative learning framework}

Suppose $K$ data owners collaboratively train a machine learning model based on a set of data $\{\bx_i, y_i\}_{i=1}^{N}$ and only one party has labels. This is a reasonable assumption in cross-organizational collaborative learning scenarios, because in reality labels (such as users' credit scores, patients' diagnosis etc) are expensive to obtain and only exist in one or few of the organizations. Suppose that the feature vector $\bx_i$ can be further decomposed into $K$ blocks $\{\bx^k_i\}_{k=1}^{K}$, where each block belongs to one owner. 
Without loss of generality, assume that the labels are located in party $K$. 
Then the collaborative training problem can be formulated as 
\begin{equation}
\label{eq:y_pred}
o_i^K = f(\theta_1,\dots,\theta_K; \bx_i^1,\dots,\bx_i^K)
\end{equation}
\begin{equation}
\label{eq:loss}
\min_{\theta} \mL(\theta; \mD) = \frac{1}{N}\sum^N_{i=1} \ell(o_i^K,y_i^K) + \lambda\sum_{k=1}^K\gamma(\theta_k)
\end{equation}
where $\theta_k$ denote the training parameters of the $k$th party; $\bTheta={[{\btheta_{1}};\dots; {\btheta_{K}}]}$; $N$ denote the total number of training samples; $f(\cdot)$ and $\gamma(\cdot)$ denote the prediction function and regularizer and $\lambda$ is the hyperparameter; Following previous work, we assume each party adopts a sub-model $G_k$ which generates local predictions, i. e, local latent representations $H_i^k$ and the final prediction is made by merging $H_k$ with an nonlinear operation, such as softmax function. That is,
\begin{equation}
\label{eq:H_k}
H_i^k = G_k(\theta_k, \bx_i^k)
\end{equation}
\begin{equation}
\label{eq:loss_H_k}
\ell(\btheta_1,\dots,\btheta_K; \mD_i)=\ell(f(\sum_{k=1}^K H_i^k),y_i^K)
\end{equation}
where $G_k$ can adopt a wide range of models such as linear and logistic regression, support vector machines, neural networks etc. Let $H_i = \sum_{k=1}^K H_i^k$, then the gradient function has the form
\begin{equation}
\label{eq:grad_H_k}
\nabla_k \ell(\btheta_1,\dots,\btheta_K; \mD_i) = \frac{\partial \ell}{\partial H_i}\frac{\partial H_i^k}{\partial \theta_k}
\end{equation}

We refer to the party having the labels as the \textit{active party}, and the rest of parties as \textit{passive parties}. In a feature-partitioned collaborative learning protocol, each passive party sends \{$H_i^k\}$ to active party, and the active party calculates $\{\frac{\partial \ell}{\partial H_i}\}$
and sends it back to passive parties for gradient update. See Algorithm \ref{alg:algo1}. We use $M$ to denote the model trained. 

\begin{algorithm}[tb]
   \caption{A feature-partitioned collaborative learning framework}
   \label{alg:algo1}
\begin{algorithmic}
\STATE {\bfseries Input:} learning rate $\eta$
\STATE {\bfseries Output:} Model parameters $\theta_1$, $\theta_2$ ... $\theta_K$
Party 1,2,\dots,$K$, initialize $\theta_1$, $\theta_2$, ... $\theta_K$. \\
\FOR{$j=1$ {\bfseries to} $N$}
\STATE Randomly sample  $S \subset [N]$  \\
\FOR{each party $k \ne\ K$ in parallel}
\STATE $\boldsymbol k$ computes and sends $\{H_i^k\}_{i \in S}$ to party $K$
\ENDFOR
\STATE party $K$ computes and sends $\{\frac{\partial \ell}{\partial H_i}\}_{i \in S}$ to all other parties;
\FOR{each party k=1,2,\dots,K in parallel}
\STATE $\boldsymbol k$ computes $\nabla_k \ell$ with equation \ref{eq:grad_H_k} 
\STATE update $\theta^{j+1}_k = \theta^{j}_k - \eta \nabla_k \ell$;
\ENDFOR
\ENDFOR
\end{algorithmic}
\end{algorithm}

\section{Backdoor Attacks}
\subsection{Threat Model} Based on Algorithm \ref{alg:algo1}, we consider the following attack model: (i) the attacker has access to the training data of one or more data parties $D_m$ and it controls the training procedure and local hyperparameters of attacked parties. (ii) The attacker can modify local updates such as training weight and gradients before sending transmitted data to other parties. (iii) The attacker does not control any benign parties' training nor does the attacker control the global communication and aggregation protocols. (iv) Contrary to traditional data poisoning attacks which focus on poisoning \textit{training data}, the attacker in collaborative learning focus on poisoning \textit{gradients and model updates} that are communicated among parties in the protocol, similar to previous works \cite{DBLP:journals/corr/SteinhardtKL17}. The fundamental difference between our threat model and the model adopted in previous works related to backdoor attack in FL (or sample-partitioned collaborative learning) is the following. First, in FL backdoor attacks, the attacker has access to the entire feature and label space information and the entire set of model parameters. However, in our scenario, the attacker only has access to a portion of the feature space of parties that it controls, and it may not have access to labels if it doesn't control the active party. It also has only a portion of the model parameters of the parties it controls; Secondly, in FL backdoor attacks, the attacker controls only a portion of the entire dataset and any model updates it sends get averaged in the server, causing a scale-down effect for the backdoor. In our setting, the backdoor will take place over the entire batch dataset in each iteration, and survive and propagate through the communication protocols. Due to these differences, the backdoor strategies for feature-partitioned scenario are fundamentally different from those for sample-partitioned scenario. 

\subsection{Adversarial Objectives} The attacker aims to train a model which achieves high performance on both the original task and the targeted backdoor task. Unlike Byzantine attacks \cite{munoz2019byzantine}, its goal is not to prevent convergence. The targeted backdoor task is to assign an attacker-chosen label to input data with a specific pattern (i. e. , a trigger). Unlike attacks in FL, the poisoned input data appear globally among all parties with distributed features, as shown in Figure \ref{fig:exp_setting}. 



Since the active party has labels, it can directly modify the label of the poisoned dataset without affecting the training protocol. During training, the active party just needs to send $\{\frac{\partial \mL^b}{\partial H_i}\} $instead of $\{\frac{\partial \mL}{\partial H_i}\} $ to successfully inject the attack, where $\mL^b$ denotes the loss function of the backdoor task.
Therefore in this work, we focus on a more challenging scenario, that is to perform attacks at passive party and defense at active party. This type of attack is more likely to appear in real-world scenarios. 
\subsection{Attacks at Passive Party}
The difficulty for a passive party to conduct training-time targeted attack is that it has no access to either labels or other passive parties' contributions at every iteration. From Algorithm \ref{alg:algo1}, the only information the passive parties obtain is the intermediate gradients $\{\frac{\partial \mL}{\partial H_i}\}$. 

One way to accomplish this attack is to infer labels from the intermediate gradients $\{\frac{\partial \mL}{\partial H_i}\}$, and this indeed can be accomplished in certain scenarios. For example, in equation \ref{eq:loss_H_k}, if $f$ represents a softmax function and $\ell$ represents a cross-entropy loss, then $g_H^i=\frac{\partial \ell}{\partial H_i}$ is a $m$-dimension vector where $m$ is the number of labels with the $j$th element being:
	$$g_{H,j}^i=\begin{cases}
			S_j,j\neq y\\ 
			S_j-1, j=y
			\end{cases}$$
where $S_j$ is the softmax function $S_j= \frac{e^{h_j}}{\sum_i e^{h_i}}$ over $H_i$. Here we abuse the notation $y$ to denote the index of the true label. It is now clear that $g_H^i$ reveals the label information because the $y$th element of $g_H^i$ will have opposite sign as compared to others. To inject the attack, the passive party only needs to replace $g_H^i$ with  
$$g_{H,j}^{b,i}=\begin{cases}
			S_j,j\neq \tau\\ 
			S_j-1, j=\tau
			\end{cases}$$
To prevent information leakage, the active party may change how the intermediate information is aggregated, e.g. by adding additional hidden layers ( \textit{a trainable active party}). If so, labels can no longer be recovered using the above approach. For such scenarios, we propose an alternative backdoor attack which will still accomplish the attacker's goal without recovering label data.  
\paragraph{Gradient-replacement backdoor}

In this approach, we assume that the passive attacker knows at least one clean sample, denoted as  $\mathcal{D}_{target}$, which has the same label as the targeted label of the backdoor task. We use this assumption because it's usually not too difficult to get the label of one sample in practice. With this assumption, the attacker records the received intermediate gradient of $\mathcal{D}_{target}$ as $g_{rec}$, and set the intermediate gradients of the poisoned samples to be $g_{rec}$, update the model parameters using $\gamma g_{rec}\frac{\partial H_i^k}{\partial \theta_k}$ according to equation \ref{eq:grad_H_k}, where $\gamma$ is an adjustable amplify ratio. See Algorithm \ref{alg:algo_gradient_poison_1}.

When the active party has trainable parameters as in Figure \ref{fig:network_structure},it will have the ability to learn the mapping between the activation value $H_i$ and the true label. Although the model parameters in the malicious party are changed by the poisoned gradient and the malicious party will output the poisoned activation value, the active party would still be able to establish a new mapping between the poisoned activation value and the true label, causing the backdoor task to fail. 
In this case, for each training sample in $\mathcal{D}_{poison}$, the malicious party can send a random-valued vector to the active party to prevent the active party from establishing the mapping between the poisoned activation value and the true label during training. In the inference time, however, the poisoned activation values, not the random vector, will be sent to the active party, thus the active party is likely to predict this sample as the same label as $\mathcal{D}_{target}$.  

\begin{algorithm}[!htb]
\caption{Gradient poisoning algorithm}
\label{alg:algo_gradient_poison_1}
\begin{algorithmic}
\STATE {\bfseries Input:}current batch of training dataset $\mathcal{X}$, original gradient matrix $\frac{\partial \ell}{\partial H}$ that the active party sends to the malicious party, dataset $\mathcal{D}_{target}$, $\mathcal{D}_{poison}$, the amplify ratio of the poisoned gradient $\gamma$
\STATE {\bfseries Output:} poisoned gradient matrix $G_{poisoned}$
\STATE $G_{poisoned}$ = $\frac{\partial \ell}{\partial H}$
\FOR{$x_i$ in $\mathcal{X}$}
\IF{$x_i$ belongs to $\mathcal{D}_{target}$}
\STATE assign the $i$th row of $\frac{\partial \ell}{\partial H}$ to a global variable $g_{rec}$
\ENDIF
\IF{$x_i$ belongs to $\mathcal{D}_{poison}$}
\STATE assign $\gamma \cdot g_{rec}$ to the $i$th row of $G_{poisoned}$
\ENDIF
\ENDFOR
\STATE return $G_{poisoned}$
\end{algorithmic}
\end{algorithm}

\begin{algorithm}[!htb]
\caption{Activation blurring algorithm}
\label{alg:algo_activation_poison}
\begin{algorithmic}
\STATE {\bfseries Input:}current batch of training dataset $\mathcal{X}$, original activation matrix $H$ that the malicious party sends to the active party, dataset $\mathcal{D}_{poison}$ 
\STATE {\bfseries Output:}blurred activation matrix $H_{blurred}$
\STATE $H_{blurred}$ = $H$
\FOR{$x_i$ in $\mathcal{X}$}
\IF{$x_i$ belongs to $\mathcal{D}_{poison}$}
\STATE generate a random value vector $h_{random}$
\STATE assign $h_{random}$ to the $i$th row of $H_{blurred}$
\ENDIF
\ENDFOR
\STATE return $H_{blurred}$
\end{algorithmic}
\end{algorithm}

\section{Defenses}
\subsection{Validation} 
Validation data sets are effective for detecting targeted misclassification attacks but are not effective when the adversarial model performs well on both the original and the backdoor task. Thus it is not considered as an overall defense strategy in our study. 

\subsection{Trainable Active Party}
\begin{figure}[!htb]
\centering
\subfigure[active party is not trainable
]{
\begin{minipage}[t]{0.45\linewidth}
\centering
\includegraphics[width=1\linewidth]{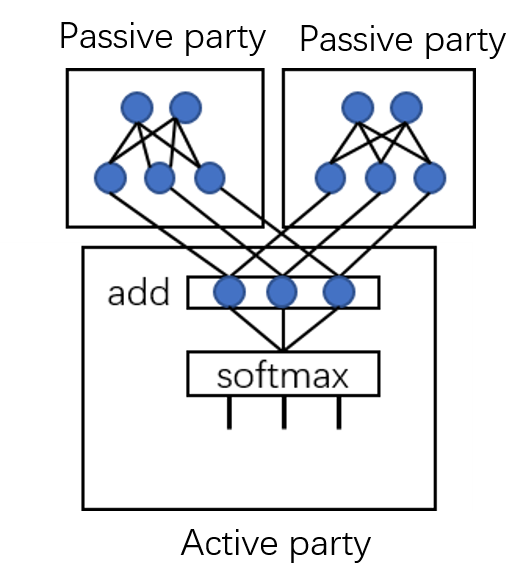}
\end{minipage}
}
\subfigure[active party is trainable]{
\begin{minipage}[t]{0.45\linewidth}
\centering
\includegraphics[width=1\linewidth]{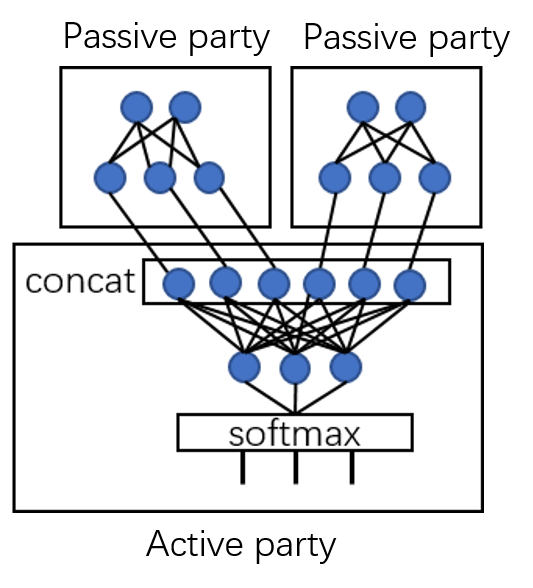}
\end{minipage}
}
\caption{two types of network structures}
\label{fig:network_structure}
\end{figure}

We introduce additional training layers at the active party for defense. The active party first concatenates the output of the passive parties then adopts a dense layer of 32 nodes before the output layer. See Figure \ref{fig:network_structure}

\subsection{Differential Privacy}
Recent works \cite{Bagdasaryan2018backdoor,xie2020dba} use differential privacy to defense the backdoor attacks in federated learning. This approach first clips the maliciously amplified gradients, then adds Gaussian or Laplacian noise to the gradients. In our experiments, we found that clipping is not as effective as adding noise to the exchanged messages. In the next section, we evaluate the effectiveness of defense at various noise level. 

\subsection{Gradient Sparsification}
Recent works \cite{DBLP:journals/corr/abs-1906-08935} have shown that gradient sparsification is an effective approach to defense attacks in federated learning. This approach prunes gradients with small magnitudes to zero. According to \cite{DBLP:journals/corr/abs-1712-01887}, the gradients can be compressed by more than 300$\times$ without losing accuracy. We use the gradient sparsification algorithm proposed in \cite{DBLP:journals/corr/abs-1712-01887}, which adopts a drop rate at each iteration during training to threshold the gradients updated. The active party sends the sparsified gradients for update but accumulates the rest of the gradients locally.


\section{Experiments}

\subsection{Models and Dataset}
\begin{figure}[!htb]
\centering
\includegraphics[width=1\linewidth]{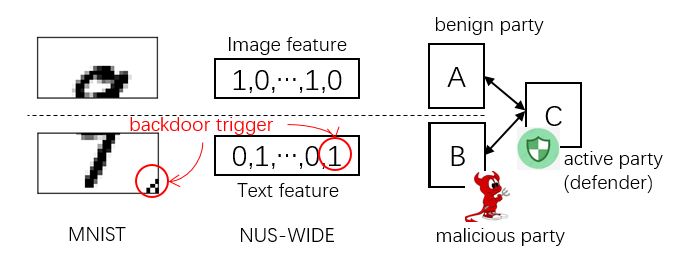}
\centering
\caption{backdoor task settings}
\label{fig:exp_setting}
\end{figure}

\paragraph{Models} There are two passive parties (party A and party B) which have different features and one active party (party C) which has the label. Party B is set to be the malicious party. Note party A and C can be the same party as well. 
See Figure \ref{fig:exp_setting}. We use tensorflow 2.0 to implement our experiments. ReLU is used as the activation function for all the nodes except for the output layer, SGD with default learning rate 0.01 is used as the optimizer and the batch size is set to 64. For all the backdoor training experiments, Algorithm \ref{alg:algo_gradient_poison_1} is used to perform the backdoor task, Algorithm \ref{alg:algo_activation_poison} is used in case of the active party is trainable, and Gaussian noise with variance of 1e-6 is used as random vectors in Algorithm \ref{alg:algo_activation_poison}. The passive party contains an input layer and an output layer, the size of the output layer is set to be 32.

\paragraph{NUS-WIDE dataset} In this dataset, each sample has 634 image features and 1000 text features which is very suitable for the feature-partitioned experiments. In this experiment, we choose ['buildings', 'grass', 'animal', 'water', 'person'] as labels in our dataset. Party A gets the image features, and party B gets text features. We choose the poisoned data samples as those whose last text feature equals 1. Table \ref{tab:nus_wide_class_distribution} shows the label distribution of $\mathcal{D}_{poison}$. We can see that most of the labels of $\mathcal{D}_{poison}$ belong to animal ($class_2$), but few of them belong to buildings ($class_0$). So we set the goal of our backdoor task to change the predicted label of $\mathcal{D}_{poison}$ to be buildings ($class_0$). The data samples that satisfy this feature are less that 1\% in both training set and testing set according to Table \ref{tab:nus_wide_class_distribution}, thus it's difficult to detect this backdoor using a validation set. 
\paragraph{MNIST dataset} In this dataset, each sample has $28\times28$ pixels. We partition the dataset so that party A gets the first 14 rows while party B gets the last 14 rows. We randomly select 600 samples from the 60000 training samples and 100 samples from the 10000 testing samples, and mark their pixels in [25,27], [27,25], [26,26] and [27,27] to be 255, similar to \cite{DBLP:journals/corr/abs-1708-06733}. 

\begin{table}[htb]
\caption{comparing number of samples of the backdoor data with the whole dataset of NUS-WIDE. }  
\begin{center}
\begin{tabular}{|l|l|l| p{2cm}|}  
\hline  
ID & Label & Train & Test \\ \hline  
0 & buildings & 3 / 6173 & 0 / 4213 \\ \hline  
1 & grass & 8 / 5981 & 8 / 4082 \\ \hline  
2 & animal & 125 / 12679 & 81 / 8296 \\ \hline  
3 & water & 11 / 12202 & 4 / 8172 \\ \hline  
4 & person & 5 / 22965 & 9 / 15237 \\  
\hline  
\end{tabular}  
\end{center}
\label{tab:nus_wide_class_distribution}
\end{table}

\begin{figure}[!htb]
\centering
\subfigure[normal(NUS-WIDE)]{
\begin{minipage}[t]{0.47\linewidth}
\centering
\includegraphics[width=1\linewidth]{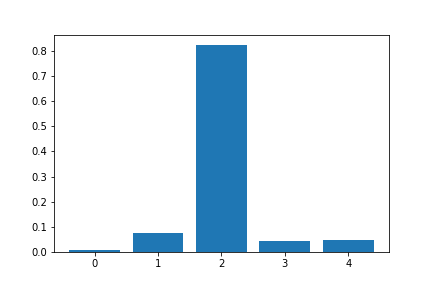}
\end{minipage}
}
\subfigure[poisoned(NUS-WIDE)]{
\begin{minipage}[t]{0.47\linewidth}
\centering
\includegraphics[width=1\linewidth]{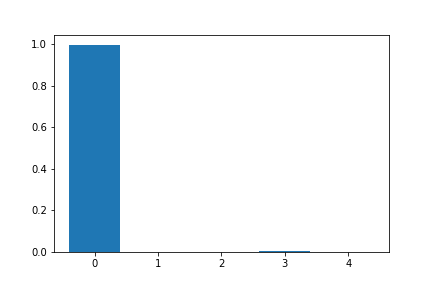}
\end{minipage}
}

\subfigure[normal(MNIST)]{
\begin{minipage}[t]{0.47\linewidth}
\centering
\includegraphics[width=1\linewidth]{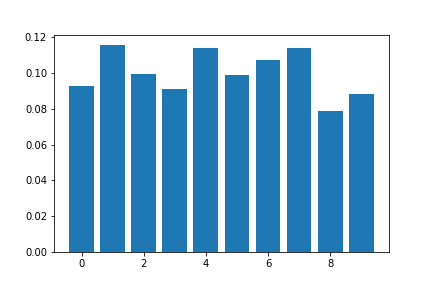}
\end{minipage}
}
\subfigure[poisoned(MNIST)]{
\begin{minipage}[t]{0.47\linewidth}
\centering
\includegraphics[width=1\linewidth]{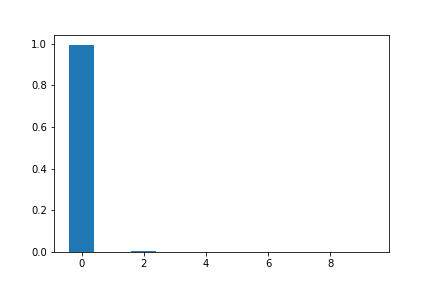}
\end{minipage}
}
\caption{average prediction score of $D_{poison}$}
\label{fig:exp2_avg_predict_score}
\end{figure}

\begin{figure}[!htb]
\centering
\subfigure[MNIST]{
\begin{minipage}[t]{0.47\linewidth}
\centering
\includegraphics[width=1\linewidth]{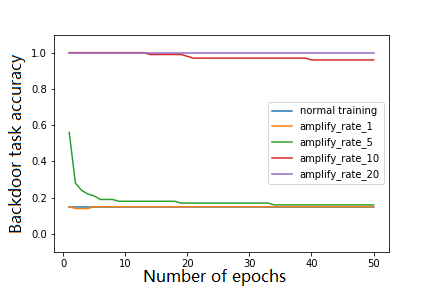}
\end{minipage}
}
\subfigure[MNIST]{
\begin{minipage}[t]{0.47\linewidth}
\centering
\includegraphics[width=1\linewidth]{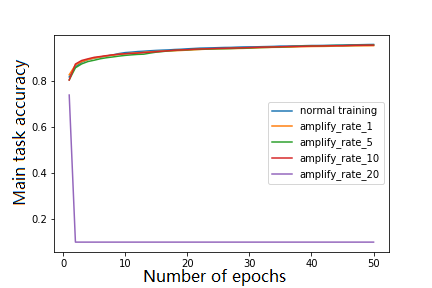}
\end{minipage}
}

\subfigure[NUS-WIDE]{
\begin{minipage}[t]{0.47\linewidth}
\centering
\includegraphics[width=1\linewidth]{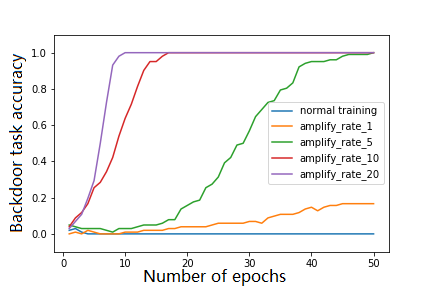}
\end{minipage}
}
\subfigure[NUS-WIDE]{
\begin{minipage}[t]{0.47\linewidth}
\centering
\includegraphics[width=1\linewidth]{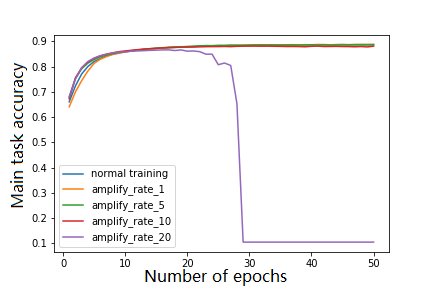}
\end{minipage}
}
\caption{comparing normal training with backdoor training when the active party is not trainable.}
\label{fig:acc_backdoor_training}
\end{figure}

\subsection{Attacks at Passive Party}

Figure \ref{fig:exp2_avg_predict_score} shows the distribution of the prediction accuracy of $\mathcal{D}_{poison}$ for normal and backdoor training. We can see that although $\mathcal{D}_{poison}$ has few samples belonging to $class_0$, the prediction score of $class_0$ is much higher than other classes, which means the backdoor task is succeeded. 

After each training epoch, we used $\mathcal{D}_{poison}$ to check the accuracy of the backdoor task and the testing set to check the accuracy of the main task. Figure \ref{fig:acc_backdoor_training} shows the results for various $\gamma$. We can see that for a reasonable range of $\gamma$, both the backdoor task and normal task maintain high accuracy, indicating a successful injection of backdoor. The gap between the normal task and the backdoor task is not obvious, so it is hard to detect this backdoor using a validation set. 

The main task and the backdoor task both work well when $\gamma$ equals $10$, thus we set $\gamma$ to be $10$ in the following experiments.

\subsection{Defenses at Active Party}


\begin{figure}[!htb]
\centering
\subfigure[MNIST]{
\begin{minipage}[t]{0.47\linewidth}
\centering
\includegraphics[width=1\linewidth]{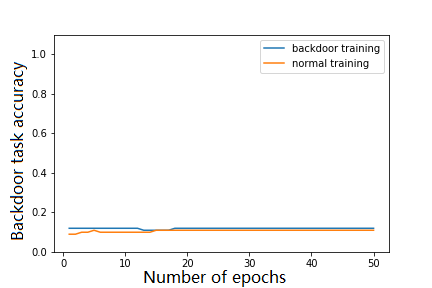}
\end{minipage}
}
\subfigure[MNIST]{
\begin{minipage}[t]{0.47\linewidth}
\centering
\includegraphics[width=1\linewidth]{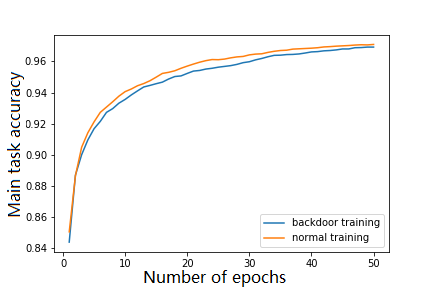}
\end{minipage}
}

\subfigure[NUS-WIDE]{
\begin{minipage}[t]{0.47\linewidth}
\centering
\includegraphics[width=1\linewidth]{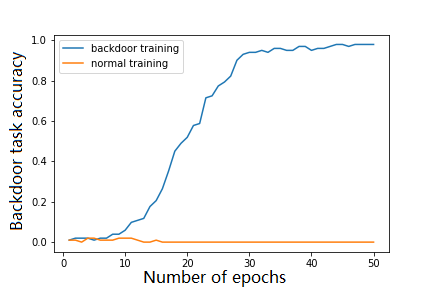}
\end{minipage}
}
\subfigure[NUS-WIDE]{
\begin{minipage}[t]{0.47\linewidth}
\centering
\includegraphics[width=1\linewidth]{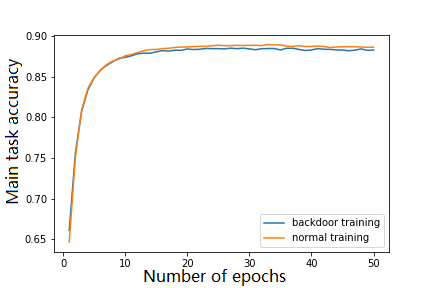}
\end{minipage}
}
\caption{comparing normal training with backdoor training when active party is trainable.}
\label{fig:acc_backdoor_training_with_hidden_layer}
\end{figure}

\begin{table}[tb]
\caption{accuracy of different network. }  
\begin{center}
\begin{tabular}{ |p{3.4cm}|p{1.2cm}|p{1.8cm}|  }
\hline
  & MNIST & NUS-WIDE \\
\hline
 untrainable active party & 95.7\% & 88.5\%\\
\hline
 trainable active party & $\mathbf{97.1\%}$ & $\mathbf{88.7\%}$ \\
\hline
\end{tabular}
\end{center}
\label{tab:accuracy_of_different_network}
\end{table}

\paragraph{Trainable active party}
Figure \ref{fig:acc_backdoor_training_with_hidden_layer} shows that the backdoor task failed in MNIST but succeeded in NUS-WIDE. Table \ref{tab:accuracy_of_different_network} shows the accuracy of the main task also improved by this network structure. Although this defense strategy failed on NUS-WIDE, we can combine it with other defense techniques. 

\begin{figure}[!htb]
\centering
\subfigure[Gaussian noise]{
\begin{minipage}[t]{0.47\linewidth}
\centering
\includegraphics[width=1\linewidth]{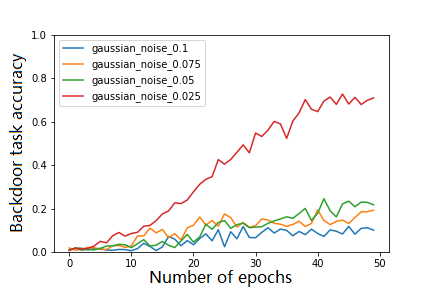}
\end{minipage}
}
\subfigure[Gaussian noise]{
\begin{minipage}[t]{0.47\linewidth}
\centering
\includegraphics[width=1\linewidth]{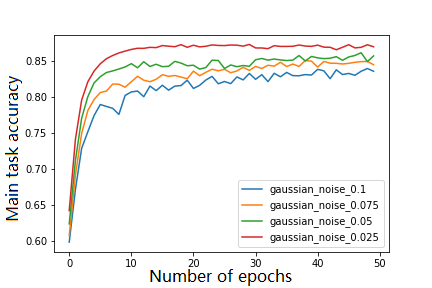}
\end{minipage}
}

\subfigure[Laplacian noise]{
\begin{minipage}[t]{0.47\linewidth}
\centering
\includegraphics[width=1\linewidth]{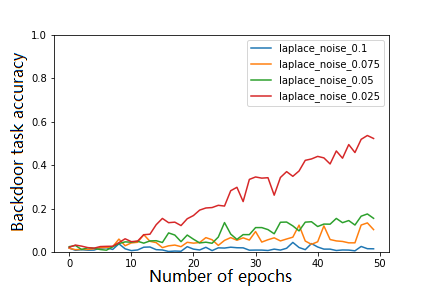}
\end{minipage}
}
\subfigure[Laplacian noise]{
\begin{minipage}[t]{0.47\linewidth}
\centering
\includegraphics[width=1\linewidth]{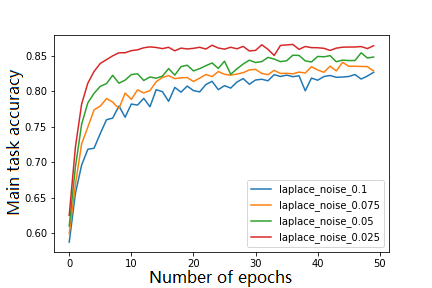}
\end{minipage}
}
\caption{comparing noisy training with different variances when active party is trainable on NUS-WIDE.}
\label{fig:acc_noisy_training_with_hidden_layer}
\end{figure}

\paragraph{Differential Privacy}
Since the backdoor task failed on MNIST with trainable active party already, we did experiments on NUS-WIDE with trainable active party. Gaussian noise and Laplacian noise with different variances were added and compared. This experiment is repeated 20 times in order to alleviate the effect of randomness. Figure \ref{fig:acc_noisy_training_with_hidden_layer} shows that noise with variances larger than 0.05 obviously alleviated the backdoor task. However, it also hurts the accuracy of the main task. 

\begin{figure}[!htb]
\centering


\subfigure[NUS-WIDE]{
\begin{minipage}[t]{0.47\linewidth}
\centering
\includegraphics[width=1\linewidth]{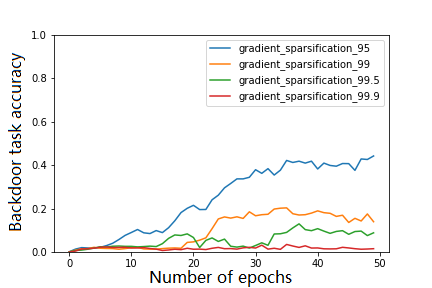}
\end{minipage}
}
\subfigure[NUS-WIDE]{
\begin{minipage}[t]{0.47\linewidth}
\centering
\includegraphics[width=1\linewidth]{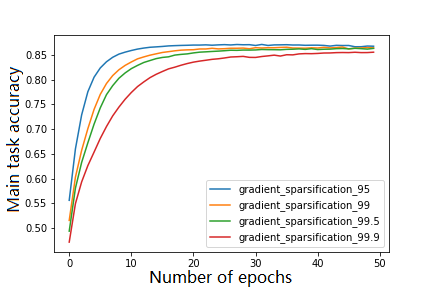}
\end{minipage}
}
\caption{comparing gradient sparsification training with different drop rate $s$. (a)-(d) is performed when the active party is not trainable, (e)-(f) is performed when the active party is trainable. }
\label{fig:acc_sparsification_training}
\end{figure}

\paragraph{Gradient Sparsification}
We then evaluate this strategy for NUS-WIDE with trainable active party where differential privacy compromises the main task accuracy.  
Figure \ref{fig:acc_sparsification_training} show the effectiveness of defenses at various sparsification levels, adjusted by the drop rate $s$. We show that gradient sparsification can mitigate the backdoor without hurting the accuracy of the main task. 


\section{Conclusion and Future Work}
In this paper we show that backdoor task can be performed by the passive party without accessing the labels in the feature-partitioned collaborative learning framework. For defense, it's recommended to use a trainable active party, which can alleviate the leakage of label information and improve the model performance. Gradient sparsification can further enhance the effectiveness of defenses without hurting the performance of main task. Future work may focus on study of additional types of attacks and defenses aiming to achieve better understanding of security in feature-partitioned collaborative learning.  

\bibliographystyle{named}
\bibliography{ref}

\end{document}